# A Dantzig Selector Approach to Temporal Difference Learning

**Matthieu Geist**                                                                                       MATTHIEU.GEIST@SUPELEC.FR
Supélec, IMS Research Group, Metz, France

**Bruno Scherrer**                                                                                       BRUNO.SCHERRER@INRIA.FR
INRIA, MAIA Project Team, Nancy, France

**Alessandro Lazaric and Mohammad Ghavamzadeh**                                                          FIRSTNAME.LASTNAME@INRIA.FR
INRIA Lille - Team SequeL, France

## Abstract

LSTD is a popular algorithm for value function approximation. Whenever the number of features is larger than the number of samples, it must be paired with some form of regularization. In particular, $\ell_1$-regularization methods tend to perform feature selection by promoting sparsity, and thus, are well-suited for high–dimensional problems. However, since LSTD is not a simple regression algorithm, but it solves a fixed–point problem, its integration with $\ell_1$-regularization is not straightforward and might come with some drawbacks (*e.g.*, the P-matrix assumption for LASSO-TD). In this paper, we introduce a novel algorithm obtained by integrating LSTD with the Dantzig Selector. We investigate the performance of the proposed algorithm and its relationship with the existing regularized approaches, and show how it addresses some of their drawbacks.

## 1. Introduction

An important problem in reinforcement learning (RL) (Sutton & Barto, 1998) is to estimate the quality of a given policy through the computation of its value function (*e.g.*, in the policy evaluation step of a policy iteration). Oftentimes, the state space is to large, and thus, approximation schemes must be used to represent the value function. Furthermore, whenever the model (reward function and probability transitions) is unknown, the best approximation should be computed using a set of sampled transitions. Many algorithms have been designed to solve this approximation problem. Among them, LSTD (Least-Squares Temporal Differences) (Bradtke & Barto, 1996) is the most popular. Using a linear parametric representation, LSTD computes the fixed-point of the Bellman operator composed with the orthogonal projection.

In many practical scenarios, the number of features of the linear approximation is much larger than the number of available samples. For example, one may want to consider a very rich function space, such that the actual value function lies in it. Unfortunately, in this case, learning is prone to overfitting. A standard approach to face this problem is to introduce some form of regularization. While LSTD has been often paired with $\ell_2$-regularization, only recently $\ell_1$-regularization (see Sec. 2.2 for a thorough review of the main $\ell_1$-regularized algorithms) has been considered to deal with high–dimensional problems. This approach is particularly appealing since $\ell_1$-regularization implicitly performs feature selection and targets sparse solutions. In particular, LASSO-TD (Kolter & Ng, 2009) can be seen as en extension of LASSO (Tibshirani, 1996) to temporal difference learning, to which it reduces when setting the discount factor to zero. However, LASSO-TD is not derived from a proper convex optimization problem, and thus, it requires some assumptions that might not hold in an off-policy setting. Although other algorithms have been proposed to overcome these drawbacks (*e.g.*, $\ell_1$-PBR by Geist & Scherrer (2011)), other disadvantages may appear.

This paper introduces a new algorithm, Dantzig-LSTD (D-LSTD for short, see Sec. 3), which extends the Dantzig Selector (DS) (Candes & Tao, 2007) to temporal difference learning. Instead of solving a fixed-point problem as in LASSO-TD, it can simply be cast as a linear program, thus allowing to use any off-the-shelf solver. Furthermore, since the underlying optimiza-tion problem is convex, it can handle off-policy learning in a principled way. Yet, when LASSO-TD is well defined, both algorithms provide similar solutions (see Prop. 2), as DS does w.r.t. LASSO. We show that for some oracle choice of the regularization factor, the D-LSTD solution converges quickly to the LSTD solution (at a rate depending only logarithmically on the number of features), as shown in Theorem 1. This new algorithm also opens some issues, namely how well is the true value function estimated and how to efficiently choose the regularization factor. These points are discussed in Sec. 4. Finally, we report some illustrative empirical results in Sec. 5.

## 2. LSTD and Related Work

A Markov reward process[1] (MRP) is a tuple $\{S, P, R, \gamma\}$, where $S$ is a finite state space, $P = (p(s'|s))_{1 \leq s, s' \leq |S|}$ is the transition matrix, $R = (r(s))_{1 \leq s \leq |S|}$ with $\|R\|_\infty \leq r_{\max}$ is the reward vector, and $\gamma$ is a discount factor. The value function $V$ is defined as the expected cumulative reward from a given state $s$, $V(s) = \mathbb{E}[\sum_{t=0}^\infty \gamma^t r_t | s_0 = s]$. It is the unique fixed-point of the Bellman operator $T : V \to R + \gamma PV$.

In many practical applications, the model of the MRP (i.e., the reward $R$ and transitions $P$) is unknown and only a set of $n$ transitions $\{(s_i, r_i, s'_i)_{1 \leq i \leq n}\}$ is available. In general, we assume that states $s_1, \ldots, s_n$ are sampled from a *sampling distribution* $\mu$ (not necessarily the stationary distribution of the MRP) and the next states $s'_1, \ldots, s'_n$ are generated according to the transition probabilities $p(\cdot|s_i)$. Whenever the state space is too large, the value function cannot be computed exactly at each state and a function approximation scheme is needed. We consider value functions $\hat{V}_\theta$ defined as a linear combination of $p$ basis functions $\phi_i(s)$, that is $\hat{V}_\theta(s) = \sum_{i=1}^p \theta_i \phi_i(s) = \theta^\top \phi(s)$. We denote by $\Phi \in \mathbb{R}^{|S| \times p}$ the feature matrix whose rows contain the feature vectors $\phi(s)^\top$ for any $s \in S$. This defines a hypothesis space $\mathcal{H} = \{\Phi\theta | \theta \in \mathbb{R}^p\}$, which contains all the value functions that can be represented by the features $\phi$. The objective is to find the function $\hat{V}_{\theta^*}$ that approximates $V$ the best.

### 2.1. LSTD

Let $\Pi_\mu$ denote the orthogonal projection onto $\mathcal{H}$ w.r.t. the sampling distribution $\mu$. If $D_\mu$ is the diagonal matrix with elements $\mu(s)$ and $M_\mu = \Phi^\top D_\mu \Phi$ is the Gram matrix, then the projection operator is $\Pi_\mu = \Phi M_\mu^{-1} \Phi^\top D_\mu$. Motivated by the fact that the value function is the fixed point of the Bellman operator $T$, the LSTD algorithm computes the fixed–point of the joint $\Pi_\mu T$ operator: $\hat{V}_{\theta^*} = \Pi_\mu T \hat{V}_{\theta^*}$. Let us define $A \in \mathbb{R}^{p \times p}$ and $b \in \mathbb{R}^p$ as $A = \Phi^\top D_\mu (I - \gamma P) \Phi$ and $b = \Phi^\top D_\mu R$. In the following we assume that $A$ and $M_\mu$ are invertible. It can be shown through simple algebra that $\hat{V}_{\theta^*}$ is the fixed-point of $\Pi_\mu T$ if and only if $\theta^*$ is the (unique) solution to $A\theta^* = b$. This relationship is particularly interesting since it shows that computing the fixed point $\Pi_\mu T$ is equivalent to solving a linear system of equations defined by $A$ and $b$.

Since $P$ and $R$ are not usually known, we have to rely on sample–based estimates. In particular, we define $\tilde{\Phi}$ (resp. $\tilde{\Phi}'$) $\in \mathbb{R}^{n \times p}$ the empirical feature matrices whose rows contain the feature vectors $\phi(s_i)^\top$ (resp $\phi(s'_i)^\top$), and $\tilde{R} \in \mathbb{R}^n$ the reward vector of row-components $r_i$. The random matrices $\tilde{A}$ and $\tilde{b}$ are then defined as $\tilde{A} = \frac{1}{n} \tilde{\Phi}^\top \Delta \tilde{\Phi}$ and $\tilde{b} = \frac{1}{n} \tilde{\Phi}^\top \tilde{R}$ with $\Delta \tilde{\Phi} = \tilde{\Phi} - \gamma \tilde{\Phi}'$. LSTD computes the solution $\theta_0$ of the sample–based linear system $\tilde{A}\theta_0 = \tilde{b}$. We notice that both $\tilde{A}$ and $\tilde{b}$ are unbiased estimators of the model–based matrices $A$ and $b$ (i.e., $\mathbb{E}[\tilde{A}] = A$ and $\mathbb{E}[\tilde{b}] = b$), thus suggesting that as the number of samples increases, the solution of LSTD $\theta_0$ converges to the model–based solution $\theta^*$. Since LSTD computes the fixed point of the joint operator $\Pi_\mu T$, then the sample–based LSTD solution can also be formulated in an equivalent form as the solution of two nested optimization problems:

$$\begin{cases} \omega_\theta = \operatorname{argmin}_\omega \|\tilde{R} + \gamma \tilde{\Phi}' \theta - \tilde{\Phi} \omega\|_2^2 \\ \theta_0 = \operatorname{argmin}_\theta \|\tilde{\Phi}\theta - \tilde{\Phi}\omega_\theta\|_2^2 \end{cases}, \quad (1)$$

where the first equation projects the image of the estimated value function $\hat{V}_\theta$ under the Bellman operator $T$ onto the hypothesis space $\mathcal{H}$, and the second one solves the related fixed-point problem.

### 2.2. Related Work

When the number of samples is close or smaller than the number of features, the matrix $\tilde{A}$ is ill–conditioned and some form of regularization should be employed to solve the LSTD problem. In this section, we review the state–of–the–art regularized LSTD algorithms.

The formulation of LSTD in Eq. 1 is particularly helpful in understanding the different regularization schemes that could be applied to LSTD. In particular, each of the minimizations relative to the operators $\Pi_\mu$ and $T$ can be regularized, thus obtaining:

$$\begin{cases} \omega_\theta = \operatorname{argmin}_\omega \|\tilde{R} + \gamma \tilde{\Phi}' \theta - \tilde{\Phi} \omega\|_2^2 + \lambda_1 \operatorname{pen}_1(\omega) \\ \theta_{\lambda_1, \lambda_2} = \operatorname{argmin}_\theta \|\tilde{\Phi}\theta - \tilde{\Phi}\omega_\theta\|_2^2 + \lambda_2 \operatorname{pen}_2(\theta) \end{cases}.$$

---

[1] This can easily be extended to Markovian decision processes that reduce to MRPs for fixed policies.

| pen$_1$ \ pen$_2$ | $\emptyset$ | $\|.\|_2$ | $\|.\|_1$ |
|---|---|---|---|
| $\emptyset$ | LSTD | ✓ | $\ell_1$-PBR |
| $\|.\|_2$ | ✓ | $\ell_{2,2}$-LSTD | $\ell_{2,1}$-LSTD |
| $\|.\|_1$ | LASSO-TD | ? | ? |

Table 1. Summary of the existing regularized LSTD algorithms (except $\ell_1$-LSTD). Checkmarks are special cases of other algorithms and question marks represent combinations that have not been yet studied in the literature.

With this formulation, all the regularization schemes for LSTD (except $\ell_1$-LSTD, which we discuss at the end of this section) can be summarized as in Tab. 1.

Ridge regression (*i.e.*, $\ell_2$-regularization) is the most common form of regularization and it simply adds a term $\lambda I$ to $\tilde{A}$. This corresponds to $\lambda_1 \text{pen}_1(\omega) = \lambda \|\omega\|_2^2$ and $\lambda_2 = 0$ and it has been generalized by Farahmand et al. (2008) with $\ell_{2,2}$-LSTD, where both penalty terms use an $\ell_2$-norm regularization. Although these approaches can help in dealing with ill–defined $\tilde{A}$ matrices, they are not specifically designed for the case of $n \ll p$, where the optimal solution is sparse. In fact, it is well-known that, unlike $\ell_1$–regularization, $\ell_2$ does not promote sparsity, and thus, it might fail when the number of samples is much smaller than the number of features.

The $\ell_1$-regularization has been introduced more recently with LASSO-TD, where the projection is replaced by an $\ell_1$-penalized projection. In this case, the nested optimization problem in Eq. 1 reduces to solving the fixed-point problem (if well defined): $\theta_{l,\lambda} = \text{argmin}_\theta \|\tilde{R} + \gamma \tilde{\Phi}' \theta_{l,\lambda} - \tilde{\Phi} \theta\|_2^2 + \lambda \|\theta\|_1$. This algorithm has been first introduced by Kolter & Ng (2009) under the name LARS-TD, where it is solved using an ad–hoc variation of the LARS algorithm (Efron et al., 2004). For LARS-TD to find a solution, $\tilde{A}$ must be a P-matrix.[2] Unfortunately, this may not be true when the sampling and stationary distributions are different (off-policy learning). Although this does not always affect the performance in practice (see some of the experiments reported in Kolter & Ng 2009), it would be desirable to remove or relax this condition. The LARS-TD idea is further developed by Johns et al. (2010), where LASSO-TD is reframed as a linear complementary problem. This allows using any off-the-shelf LCP solver (notably some of them allow warm-starts, which may be of interest in a policy iteration context), but the P-matrix condition is still required, since it is inherent to the optimization problem and not to how it is actually solved. Finally, the theoretical properties of LASSO-TD were analyzed by Ghavamzadeh et al.

---

[2] A P-matrix is a matrix that all its principal minors are positive (generalizing positive definite matrices).

(2011), who provided prediction error bounds in the on-policy fixed design setting (*i.e.*, the performance is evaluated on the points in the training set). In particular, they show that, similarly to LASSO in regression, the prediction error depends on the sparsity of the projection of the value function (*i.e.*, the $\ell_0$-norm of the $\theta$ parameter of $\Pi_\mu V$), and it scales only logarithmically with the number of features. This implies that even if the dimensionality of $\mathcal{H}$ is much larger than the number of samples, the LASSO-TD accurately approximates the true value function in the on–policy setting. In order to alleviate the P-matrix problem, the $\ell_1$-PBR (Projected Bellman residual) (Geist & Scherrer, 2011) and the $\ell_{2,1}$-LSTD (Hoffman et al., 2011) algorithms have been proposed. The idea is to place the $\ell_1$-regularization term in the fixed-point equation instead of the projection equation. This corresponds to adding an $\ell_1$-penalty term to the projected Bellman residual minimization (writing $\tilde{\Pi}$ the empirical projection and $\tilde{T}$ the sampled Bellman operator): $\theta_{\text{pbr},\lambda} = \text{argmin}_\theta \|\tilde{\Pi}(\tilde{\Phi}\theta - \tilde{T}(\tilde{\Phi}\theta))\|^2 + \lambda \|\theta\|_1$. Since this is a convex optimization problem, there is no problem for $\tilde{A}$ not being a P-matrix, and off-the-shelf LASSO solvers can be used. However, this comes at the cost of a high computational cost if $n \ll p$ (notably in the computation the empirical projection, which could be as bad as $O(p^3)$), and there is no theoretical analysis.

Finally, a novel approach has been introduced by Pires (2011). The idea is to consider the linear system formulation of LSTD (*i.e.*, $A\theta = b$) and to add an $\ell_1$-penalty term to it: $\theta_{1,\lambda} = \text{argmin}_\theta \|\tilde{A}\theta - \tilde{b}\|_2^2 + \lambda \|\theta\|_1$. We refer to this algorithm as $\ell_1$-LSTD. Being defined as a proper convex optimization problem, it does not have theoretical problems in the off-policy setting and any standard solver can be used. Notice that for $\gamma = 0$, $\ell_1$-LSTD does not reduce to a known algorithm.

## 3. Dantzig-LSTD

The Dantzig-LSTD (D-LSTD for short) algorithm that we propose in this paper returns an estimate $\theta_{d,\lambda}$ (*i.e.*, a value function $V_{\theta_{d,\lambda}}$) with a low $\ell_1$-norm under the constraint that the Bellman residual ($\tilde{R} + \gamma \tilde{\Phi}' \theta - \tilde{\Phi}\theta$), namely the correlated Bellman residual ($\tilde{\Phi}^\top(\tilde{R} + \gamma\tilde{\Phi}'\theta - \tilde{\Phi}\theta) = \tilde{b} - \tilde{A}\theta$), is smaller than a parameter $\lambda$. Formally, D-LSTD solves:

$$\theta_{d,\lambda} = \underset{\theta \in \mathbb{R}^p}{\text{argmin}} \|\theta\|_1 \quad \text{subject to } \|\tilde{A}\theta - \tilde{b}\|_\infty \leq \lambda. \quad (2)$$

This optimization problem is convex and can be easily recast as a linear program (LP):

$$\min_{u,\theta \in \mathbb{R}^p} \mathbf{1}^\top u \quad \text{subject to } \begin{cases} -u \leq \theta \leq u \\ -\lambda \mathbf{1} \leq \tilde{A}\theta - \tilde{b} \leq \lambda \mathbf{1} \end{cases}.$$

This algorithm is closely related to DS (Candes & Tao, 2007), to which it reduces when $\gamma = 0$. Being a convex optimization problem, it does not require $\tilde{A}$ to be a P-matrix and it can be solved using any LP solver (notably the efficient primal–dual interior point method of Candes & Tao 2007, which makes use of the Woodbury matrix identity when $n \ll p$).

### 3.1. A Finite Sample Analysis

In this section we study how well the D-LSTD solution $\theta_{d,\lambda}$ compares to $\theta^*$, i.e., the model–based LSTD solution satisfying $A\theta^* = b$. The analysis follows similar steps as in Pires (2011) for $\ell_1$-LSTD. In the following, we use the assumption that the samples are generated i.i.d. from an arbitrary sampling distribution $\mu$. We leave as future work the extension to Markov design (i.e., when the samples are generated from a single trajectory of the policy under evaluation).

**Theorem 1.** *Let $B_{\infty,\phi} = \max_{s \in S} \|\phi(s)\|_\infty$, the D-LSTD solution $\theta_{d,\lambda}$ (Eq. 2) satisfies*

$$\inf_\lambda \|A\theta_{d,\lambda} - b\|_\infty \leq \qquad (3)$$
$$2\left(\|\theta^*\|_1 (1+\gamma) B_{\infty,\phi} + r_{max}\right) B_{\infty,\phi} \sqrt{\frac{4}{n} \ln \frac{8p}{\delta}},$$

*with probability at least $1 - \delta$.*

*Proof. (sketch)* We first need a concentration result for the $\ell_\infty$-norm. Let $x_1, \ldots, x_n$ be i.i.d. random vectors with mean $\bar{x} \in \mathbb{R}^d$ and bounded by $\|x_i\|_\infty \leq B$. Using Hoeffding inequality and a union bound, it is easy to show that with probability greater that $1 - \delta$, one has $\|\frac{1}{n} \sum_{i=1}^n x_i - \bar{x}\|_\infty \leq B\sqrt{\frac{2}{n} \ln \frac{2d}{\delta}}$. Let $\Delta_{A,\max} = \|A - \tilde{A}\|_{\max}$ (entrywise max norm) and $\Delta_{b,\max} = \|b - \tilde{b}\|_\infty$. We have the following consistency inequality: $\|A\theta\|_\infty \leq \|A\|_{\max} \|\theta\|_1$. Combined with the triangle inequality, this gives: $\left| \|A\theta - b\|_\infty - \|\tilde{A}\theta - \tilde{b}\|_\infty \right| \leq \Delta_{A,\max} \|\theta\|_1 + \Delta_{b,\max}$. Let us choose $\lambda = \Delta_{A,\max} \|\theta^*\|_1 + \Delta_{b,\max}$. The previous inequality implies that $\|\tilde{A}\theta^* - \tilde{b}\|_\infty \leq \lambda$ (recall that $A\theta^* = b$). Combined with the fact that $\theta_{d,\lambda}$ minimizes Eq. 2, we have that $\|\theta_{d,\lambda}\|_1 \leq \|\theta^*\|_1$. Combining the previous results, we obtain $\|A\theta_{d,\lambda} - b\|_\infty \leq 2\Delta_{A,\max} \|\theta^*\|_1 + 2\Delta_{b,\max}$. The concentration result for $\|.\|_\infty$ can be used to bound $\Delta_{A,\max}$ and $\Delta_{b,\max}$, which gives the stated result, using the fact that $\|\phi(s_i)(\phi(s_i) - \gamma\phi(s_i'))^T\|_{\max} \leq B_{\infty,\phi}^2 (1+\gamma)$ and that $\|\phi(s_i) r_i\|_\infty \leq B_{\infty,\phi} r_{\max}$. □

Since the algorithm is specifically designed for the high–dimensional setting ($n \ll p$), it is critical to study the dependency of the performance on $n$ and $p$. Up to constant terms, the previous bound can be written as

$$\inf_\lambda \|A\theta_{d,\lambda} - b\|_\infty \leq O\left(\|\theta^*\|_1 \sqrt{\frac{1}{n} \ln \frac{p}{\delta}}\right).$$

First we notice that as the number of samples increases, the error of $\theta_{d,\lambda}$ tends to zero, thus implying that it matches the performance of the model–based LSTD solution $\theta^*$. Furthermore, the dependency on the number of features $p$ is just logarithmic, while the $\ell_1$-norm of $\theta^*$ is assumed to be small whenever the solution is sparse. This suggests that D-LSTD could work well even in the case $n \ll p$ whenever the problem admits a sparse LSTD solution. Finally, we also notice that there is no specific assumption regarding the learning setting, except that $A$ should be invertible. This is particularly important because it means that, unlike most of the other results available for LSTD (e.g., see Ghavamzadeh et al. 2011), this result holds also in the off-policy setting. The main drawback of this analysis is that it holds for an oracle choice of $\lambda$. We postpone a discussion about how to choose the regularizer in practice to Sec. 4.

### 3.2. Comparison to Other Algorithms

Similar to $\ell_1$-PBR and $\ell_{2,1}$-LSTD, D-LSTD is based on a well-defined standard convex optimization problem, which does not require $\tilde{A}$ to be a P-matrix (unlike LASSO-TD) and that can be solved using any off-the-shelf solvers. Nonetheless, D-LSTD has only one meta-parameter (instead of two), and in general, it has a smaller computational cost w.r.t. solving the nested optimization problems of $\ell_1$-PBR and $\ell_{2,1}$-LSTD.

D-LSTD is also related to LASSO-TD:

**Proposition 2.** *The LASSO-TD solution $\theta_{l,\lambda}$ (if it exists) satisfies the D-LSTD constraints:*

$$\|\tilde{A}\theta_{l,\lambda} - \tilde{b}\|_\infty \leq \lambda.$$

*Proof.* The optimality conditions of LASSO-TD can be obtained by ensuring that 0 belongs to the subgradient of $\frac{1}{2}\|\tilde{\Phi}\theta - (\tilde{R} + \gamma\tilde{\Phi}'\theta_{l,\lambda})\|_2^2 + \lambda\|\theta\|_1$ and then substituting $\theta$ by $\theta_{l,\lambda}$ (Kolter & Ng, 2009). This notably implies that for all $1 \leq i \leq p$, we have $-\lambda \leq (\tilde{b} - \tilde{A}\theta_{l,\lambda})_i \leq \lambda$, which is the stated result. □

Therefore, D-LSTD and LASSO-TD satisfy the same constraints, but $\|\theta_{l,\lambda}\|_1 \geq \|\theta_{d,\lambda}\|_1$, thus suggesting a more sparse solution. This is not surprising, since D-LSTD relates to LASSO-TD in a similar way as DS does to LASSO (Bickel et al., 2009). However, thanks to its definition as a convex optimization problem, D-LSTD avoids the main drawbacks of LASSO-TD (notably the P-matrix requirement).

Similar to $\ell_1$-LSTD, D-LSTD is built on the linear system of equations formulation of LSTD. Both approaches relax the condition $\tilde{A}\theta = \tilde{b}$ (using an $\ell_2$-norm of the error for $\ell_1$-LSTD and an $\ell_\infty$-norm for D-LSTD) while penalizing model complexity through the $\ell_1$-norm of the parameter vector. Both algorithms have the same advantages compared to LASSO-TD and to $\ell_1$-PBR/$\ell_{2,1}$-LSTD. Their main difference lies in their convergence rate. A result similar to Theorem 1 exists for $\ell_1$-LSTD (Pires, 2011):

$$\inf_\lambda \|A\theta_{1,\lambda} - b\|_2 \leq O\left(\|\theta^*\|_1 \sqrt{\frac{p^2}{n} \ln \frac{1}{\delta}}\right).$$

Although controlling the $\ell_2$-norm (in $\ell_1$-LSTD) may be harder than the $\ell_\infty$-norm (as in D-LSTD), $\ell_1$-LSTD has a very poor dependency on $p$, which makes the bound not informative as $n \ll p$. On the other hand, D-LSTD just has a logarithmic dependency on $p$.

## 4. Discussion

In this section, we discuss how the error $\|A\theta - b\|$ relates to the value function prediction error and how to choose the regularizer $\lambda$ in practice.

### 4.1. From the Parameters to the Value

Similar to Yu & Bertsekas (2010), we can link $V - \hat{V}_\theta$ to $A\theta - b$ as in the next theorem.

**Theorem 3.** *For any $\hat{V}_\theta = \Phi\theta$, we have the component–wise equality:*

$V - \hat{V}_\theta = (I - \gamma\Pi_\mu P)^{-1}((V - \Pi_\mu V) + \Phi M_\mu^{-1}(A\hat{\theta} - b)).$

*Proof.* Recall that $V = TV$ (for the true value function) and that $\hat{V}_\theta = \Pi_\mu \hat{V}_\theta$ (for an estimate $\hat{V}_\theta = \Phi\theta$ belonging to the hypothesis space). We have that:

$V - \Pi_\mu V = V - \Pi_\mu TV - (\hat{V}_\theta - \Pi_\mu T\hat{V}_\theta) + (\hat{V}_\theta - \Pi_\mu T\hat{V}_\theta)$
$\quad = (I - \gamma\Pi_\mu P)(V - \hat{V}_\theta) + \Pi_\mu(\hat{V}_\theta - T\hat{V}_\theta),$
$V - \hat{V}_\theta = (I - \gamma\Pi_\mu P)^{-1}((V - \Pi_\mu V) + \Pi_\mu(T\hat{V}_\theta - \hat{V}_\theta)).$

Note that $\Pi_\mu(T\hat{V}_\theta - \hat{V}_\theta) = \Phi M_\mu^{-1}(b - A\theta)$ gives the result. □

In order to have the final prediction error, we apply the $\ell_\infty$-norm to Theorem 3. Let $L_\mu^\phi = \max_s \|M_\mu^{-1}\phi(s)\|_1$, using Theorem 1, we obtain

$$\inf_\lambda \|V - \hat{V}_{\theta_{d,\lambda}}\|_\infty \leq \|(I - \gamma\Pi_\mu P)^{-1}\|_\infty \times$$
$$\left(\|V - \Pi_\mu V\|_\infty + O\left(\|\theta^*\|_1 L_\mu^\phi \sqrt{\frac{1}{n} \ln \frac{p}{\delta}}\right)\right).$$

In general, the previous expression cannot be simplified any further. Nonetheless, under a high–dimensional assumption $\Pi_\mu P = P$ and $\Pi_\mu R = R$. Therefore, the hypothesis space $\mathcal{H}$ is stable by the Bellman operator $T$ and $V \in \mathcal{H}$. In this case, we have $\|V - \Pi_\mu V\|_\infty = 0$ and it can be shown that $\|(I - \gamma\Pi_\mu P)^{-1}\|_\infty = \frac{1}{1-\gamma}$. Thus, we obtain the bound (valid also in the off-policy case):

$$\inf_\lambda \|V - \hat{V}_{\theta_{d,\lambda}}\|_\infty \leq O\left(\frac{\|\theta^*\|_1 L_\mu^\phi}{1-\gamma} \sqrt{\frac{1}{n} \ln \frac{p}{\delta}}\right).$$

The main critical term in this bound is $L_\mu^\phi$, which might hide a dependency on the number of features $p$. In fact, although the specific value of $L_\mu^\phi$ depends on the feature space, it is possible to find cases when it grows as $\sqrt{p}$ (consider an orthonormal basis), thus potentially neutralizing the low dependency on $p$ in Theorem 1. It is an open question whether this dependency on $p$ is intrinsic to the algorithm or is an artifact of the proof. In fact, if $\theta_{d,\lambda}$ solves the linear system of equations accurately, then we expect that the corresponding function $\hat{V}_{\theta_{d,\lambda}}$ performs almost as well as the model–based solution $\hat{V}_{\theta^*}$. The experiments of Section 5 seem to confirm this conjecture.

### 4.2. Cross Validation

The result of Theorem 1 holds for an oracle value of $\lambda$. In practice, the choice of $\lambda$ can only be directed by the available data. This issue is of great practical importance, though not often discussed in the RL literature (especially for $\ell_1$-penalized LSTD variations). In supervised learning, algorithms minimize a risk being defined as the (empirical) expectation of some loss function. Cross-validation consists in using an independent sample to estimate the true risk function, and the meta-parameter is selected as the one minimizing the estimated true risk. However, for value function estimation, there is no such risk, and cross-validation cannot be used. A general model selection method has been derived for value function estimation by Farahmand & Szepesvári (2011). However, we may devise an ad–hoc (and simple) solution for D-LSTD. Since D-LSTD is defined as a proper convex optimization problem (which reduces to a supervised learning problem when $\gamma$ tends to 0), one may be tempted to use standard cross-validation. Unfortunately, this is not directly possible. Indeed, $\|\tilde{A}\theta - \tilde{b}\|_\infty$ is the loss ($\|.\|_\infty$) of an empirical average ($\tilde{A}$ and $\tilde{b}$) rather than the empirical expectation of a loss. However, we can still consider some heuristics. Assume that we want to estimate $\|A\theta - b\|_\infty$ for some fixed parameter vector $\theta$. Let $\tilde{A}$, $\tilde{b}$ be unbiased estimates of $A$, $b$, then $\|A\theta - b\|_\infty \le \mathbb{E}[\|\tilde{A}\theta - \tilde{b}\|_\infty]$ (Jensen's inequality). Thus, given an independent set of samples, we have access to an unbiased estimate of an upper–bound of $\|A\theta - b\|_\infty$. Based on this evidence, we propose a K-fold cross-validation-based heuristic for D-LSTD. Assume that the training set is split in $K$ folds $\mathcal{F}_k$. Let $\theta_{d,\lambda}^{(-k)}$ denote the estimate trained without $\mathcal{F}_k$, and $\tilde{A}_{\mathcal{F}_k}$ and $\tilde{b}_{\mathcal{F}_k}$ be the quantities computed with only the samples in $\mathcal{F}_k$. A heuristic is to choose the $\lambda$ that minimizes

$$J_1(\lambda) = \frac{1}{K} \sum_{i=1}^{K} \|\tilde{A}_{\mathcal{F}_k} \theta_{d,\lambda}^{(-k)} - \tilde{b}_{\mathcal{F}_k}\|_\infty. \quad (4)$$

However, since we are interested in the case $n \ll p$ and the estimate $\tilde{A}_{\mathcal{F}_k}$ is computed with $\frac{n}{K}$ samples, it may have a high variance. An alternative (which we empirically found to be more efficient), at the cost of adding some bias, is to choose $\lambda$ by minimizing

$$J_2(\lambda) = \frac{1}{K} \sum_{i=1}^{K} \|\tilde{A}\theta_{d,\lambda}^{(-k)} - \tilde{b}\|_\infty. \quad (5)$$

A similar heuristic can be devised for $\ell_1$-LSTD. Although the previous heuristic worked well in our experiments, it does not have any theoretical guarantees. A different model selection strategy has been devised for $\ell_1$-LSTD by Pires (2011). It consists in choosing $\hat{\lambda} = \operatorname{argmin}_{[a,b]} \|\tilde{A}\theta_{1,\lambda} - \tilde{b}\|_2^2 + \lambda'\|\theta_{1,\lambda}\|_1$ with $[a,b]$ an exponential grid and $\lambda'$ can be computed from data (no oracle choice). This does not require splitting the learning set while ensuring a bound for $\|A\theta_{1,\hat{\lambda}} - b\|_2$. We leave the adaptation of this model selection strategy to D-LSTD for future work.

## 5. Illustration and Experiment

Sec. 5.1 presents an example that shows D-LSTD alleviates the potential problem of off-policy learning. Sec. 5.2 reports a more complex corrupted chain illustrating the case of $n \ll p$, in an on- and off-policy setting, and studies (heuristic) cross-validation.

### 5.1. A Pathological MDP

We consider a simple two-state MDP (e.g., see Kolter & Ng 2009). The transition matrix and reward vector are $P = \begin{pmatrix} 0 & 1 \\ 0 & 1 \end{pmatrix}$ and $R = \begin{pmatrix} 0 & -1 \end{pmatrix}^\top$. The optimal value function is therefore $V = \frac{-1}{1-\gamma}\begin{pmatrix} \gamma & 1 \end{pmatrix}^\top$ with $\gamma$ the discount factor. Let us consider the one-feature approximation $\Phi = \begin{pmatrix} 1 & 2 \end{pmatrix}^\top$. We compare the (asymptotic) regularization paths of LASSO-TD (Kolter & Ng, 2009), $\ell_1$-LSTD and D-LSTD, in the on-policy and off-policy cases (where LASSO-TD fails).

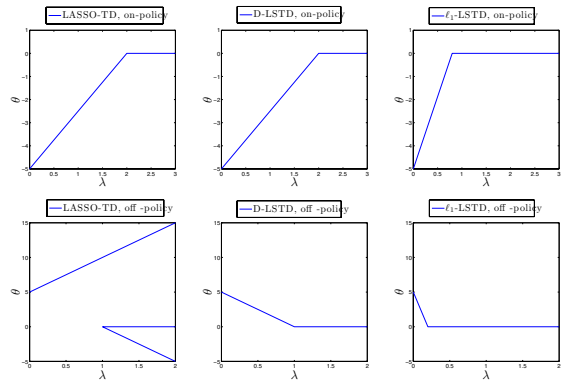

Figure 1. Two-state MDP, regularization paths.

**On-policy Case.** In the on-policy case, the sampling distribution is $\mu^\top = \begin{pmatrix} 0 & 1 \end{pmatrix}$. The regularization paths for each algorithm can be computed easily by solving analytically the optimality conditions (there is only one parameter) and they are reported in Fig. 1, top panels. LASSO-TD and D-LSTD have the same regularization path. This was expected, as there is only one parameter, but this is not true in general (recall that LASSO-TD and D-LSTD inherit the same differences as LASSO and DS).

**Off-policy Case.** Let us now consider the uniform distribution $\mu^\top = \begin{pmatrix} \frac{1}{2} & \frac{1}{2} \end{pmatrix}$. For $\gamma > \frac{5}{6}$, $A$ is not a P-matrix and LASSO-TD does not have a unique solution, nor a piecewise linear regularization path. Paths are shown on Fig. 1, bottom panels. The $\ell_1$-LSTD's path is still well-defined. LASSO-TD has more than one solution. The interesting fact here is that D-LSTD's path is well-defined, there is always a unique solution, and the path is piecewise linear. Note that both in the on- and off-policy cases all the algorithms provide the LSTD solution for $\lambda = 0$.

### 5.2. Corrupted Chain

We consider the same chain problem as in Kolter & Ng (2009) and Hoffman et al. (2011). The state $\mathbf{s}$ has $\bar{s} + 1$ components $\mathbf{s}^i$. The first one is an integer ($\mathbf{s}^1 \in \{1 \ldots 20\}$) that evolves according to a 20-state, 2-action MDP (states are connected by a chain, the action chooses the direction, and the probability of success is 0.9). All other state components are random Gaussian noises $\mathbf{s}_t^{i+1} \sim \mathcal{N}(0,1)$. The reward is $+1$ if $\mathbf{s}_t^1 = 1$ or 20. The feature vector $\phi(\mathbf{s}) \in \mathbb{R}^{\bar{s}+6}$ consists of an intercept (constant function), 5 radial basis functions corresponding to the first state component, and $\bar{s}$ identity functions corresponding to the irrelevant components: $\phi(\mathbf{s}) = \begin{pmatrix} 1 & \text{RBF}_1(\mathbf{s}^1) \ldots \text{RBF}_5(\mathbf{s}^1) & \mathbf{s}^2 \ldots \mathbf{s}^{\bar{s}+1} \end{pmatrix}^\top$. We compare LASSO-TD (with its LARS-like implementation), $\ell_1$-LSTD, and D-LSTD (for which we used $\ell_1$-magic (Romberg, 2005)).[3] We standardize the data by removing the intercept, centering the observations, centering and standardizing the features $\tilde{\Phi}$, and applying the same transformation (computed from $\tilde{\Phi}$) to $\tilde{\Phi}'$. The intercept can be computed analytically, it is the mean Bellman error (without regularization, this allows recovering the LSTD solution). We also consider $\ell_{2,\emptyset}$-LSTD, i.e., the standard $\ell_2$-penalized LSTD.

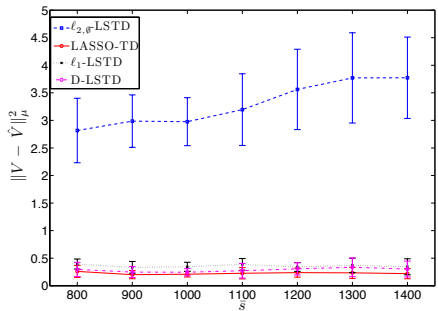

Figure 2. Corrupted chain problem – on-policy setting.

**On-policy Evaluation.** We first study the on-policy problem. The evaluated policy is the optimal one (going left if $\mathbf{s}^1 \leq 10$, and right otherwise). We sample 400 transitions (20 trajectories of length 20 started randomly on $\{1\ldots 20\}$) and vary the number $\bar{s}$ of irrelevant features between 800 and 1400. Results are presented in Fig. 2, averaged over 20 independent runs. For LARS-TD, we computed the whole regularization path (at least until too many features are added) and trained the other algorithms for a set of regularization parameters (logarithmically spaced between $10^{-3}$ and 10). Each time, we report the best prediction error (on 500 test points, such that the first state component is uniformly sampled from $\{1\ldots 20\}$), computed w.r.t. the true value function (therefore, this is an oracle choice). All $\ell_1$-penalized approaches perform significantly better than the $\ell_2$-penalization ones, showing that their performance have only a very mild dependency on the dimensionality $p$ (as predicted by Theorem 1). Among them, LASSO-TD seems to be consistently better, followed closely by D-LSTD and $\ell_1$-LSTD. For LASSO-TD and $\ell_{2,\emptyset}$-LSTD, these results are consistent with those published by Hoffman et al. (2011). Notice that there was more choice of regularization parameters for LASSO-TD, as the whole regularization path was computed. This may explain the better results of LASSO-TD compared to D-LSTD.

---
[3]We also considered $\ell_1$-PBR/$\ell_{2,1}$-LSTD. Results are not reported for the sake of clarity, but they behave like $\ell_1$-LSTD, so worse than LASSO-TD/D-LSTD.

| Algorithm | Error (mean ± std) |
|---|---|
| $\ell_{2,\emptyset}$-LSTD (oracle) | $2.82 \pm 0.58$ |
| LASSO-TD (oracle) | $0.26 \pm 0.10$ |
| $\ell_1$-LSTD (cv, $J_1/J_2$) | $4.14 \pm 0.84/0.34 \pm 0.12$ |
| D-LSTD (cv, $J_1/J_2$) | $0.65 \pm 0.18/0.23 \pm 0.11$ |

Table 2. Cross-validation.

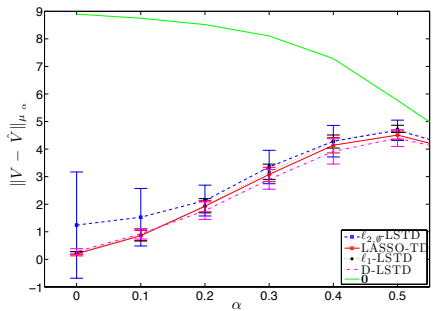

Figure 3. Corrupted chain problem – off-policy setting.

**Heuristic Cross-validation.** All results of Fig. 2 require an oracle to choose the right regularization parameter. This is not practical in a real setting. As explained in Sec. 4, $\ell_1$-LSTD and D-LSTD can benefit from a heuristic cross-validation scheme. We tested K-fold cross-validation (with $K = 5$) on this problem, with the schemes $J_1$ (Eq. 4) and $J_2$ (Eq. 5) for $n = 400$ training samples and $\bar{s} = 800$ irrelevant features (results averaged over 20 independent runs). Results are reported in Tab. 2. The error is computed as before, but here it is not used to choose the regularization parameter. The results for $J_1$ are quite bad, probably due to the high variance of the related estimator (still, for D-LSTD, the right regularization parameter is often chosen, apart from a few outliers). The $J_2$ scheme is much better, comparable to the oracle scheme (see Fig. 2). Comparing the results of the $J_2$ heuristic using a Behrens-Fisher t-test, $\ell_1$-LSTD and LASSO-TD are different (5% risk), but not D-LSTD and LASSO-TD.

**Off-policy Evaluation.** Here we test the off-policy evaluation problem. Let $\pi_{\text{opt}}$ be the optimal policy (going left if $\mathbf{s}^1 \leq 10$ and right otherwise) and $\pi_{\text{worst}} = 1 - \pi_{\text{opt}}$ (going right if $\mathbf{s}^1 \leq 10$ and left otherwise). We define $\pi_\alpha = (1-\alpha)\pi_{\text{opt}} + \alpha\pi_{\text{worst}}$, with $\alpha \in [0, \frac{1}{2}]$. Let also $\mu_\alpha$ be the corresponding stationary distribution and recall $V$ is the true value function. We consider the same problem as before, with $\bar{s} = 800$. For values of $\alpha$ varying from 0 to 0.5, we sample $n = 400$ chain states according to $\mu_\alpha$ as well as the associated transitions according to the optimal policy. The regularization parameter is chosen to minimize the error between the true value function and the estimated one on the training set (thus, an oracle-like selection procedure), for all algorithms. Results are averaged over 50 independent runs. Fig. 3 shows the error $\|\hat{V}_\alpha - V\|_{\mu_\alpha}$ as a function of $\alpha$. The term **0** corresponds to the zero prediction, that is $\|V\|_{\mu_\alpha}$. In all cases, D-LSTD seems to be slightly better than the others, and things get worse as going away from the stationary distribution (as $\alpha$ increases). In no case LASSO-TD seems to suffer from off-policy learning, suggesting that in this case the $P$-matrix condition is satisfied. Also, the difference between $\ell_2$- and $\ell_1$-schemes decreases as $\alpha$ increases. An $\ell_1$-schemes may help when there are much more features than samples, but there is little to do when the mismatch between distributions increases. Even if not reported, all approaches performed equally bad when $\alpha$ tends to one, since there is no more valuable information in the data.

## 6. Conclusion

In this paper, we introduced the Dantzig-LSTD algorithm with the objective of removing the drawbacks of existing $\ell_1$-schemes for temporal difference learning. Since D-LSTD is defined as a standard linear program, it does not require $\tilde{A}$ to be a P-matrix and can be computed using any LP solver. The D-LSTD estimate is a good approximation of the asymptotic LSTD solution in the sense of Theorem 1. It is also close to the LASSO-TD estimate (whenever well defined) in the sense of Prop. 2. In fact, D-LSTD inherits the same difference that the Dantzig selector has w.r.t. LASSO. Also, our preliminary experiments show that D-LSTD performs at least as well as LASSO-TD.

There are still a number of issues that need further investigation. As discussed in Sec. 4, when moving from the linear system of equations to the prediction error, an additional dependency on the number of features seems to appear. To which extent this dependency is an artifact of the proof or a characteristic of the algorithm is not fully clear yet. As for the choice of the regularization parameter, we plan to adapt the model selection scheme of $\ell_1$-LSTD (Pires, 2011) to D-LSTD and test it. Finally, we plan to test D-LSTD in control schemes (*i.e.*, policy iteration).

**Acknowledgments** The first author thanks the Région Lorraine for financial support. The third and fourth authors would like to thank French National Research Agency (ANR) under project LAMPADA $n°$ ANR-09-EMER-007, European Community's Seventh Framework Programme (FP7/2007-2013) under grant agreement $n°$ 231495, and PASCAL2 European Network of Excellence for supporting their research.